\documentclass[letterpaper]{article} 
\usepackage{aaai25}  
\usepackage{times}  
\usepackage{helvet}  
\usepackage{courier}  
\usepackage[hyphens]{url}  
\usepackage{graphicx} 
\usepackage{natbib}  
\usepackage{caption} 
\usepackage{algorithm}
\usepackage{algorithmic}
\usepackage{xcolor}         
\usepackage{booktabs} 
\usepackage{multirow}
\usepackage{amssymb}
\usepackage{dsfont}
\newcommand{\jy}{\textcolor[rgb]{0.0,0,0}}
\newcommand{\jynew}{\textcolor[rgb]{0.0,0,0}}
\newcommand{\jynn}{\textcolor[rgb]{0.0,0,0}}
\newcommand{\ryn}{\textcolor[rgb]{0.0,0,0}}
\newcommand{\rynn}{\textcolor[rgb]{0,0,0}}

\newcommand{\zyjnew}{\textcolor[rgb]{0.0,0,0}}
\newcommand{\zyj}{\textcolor[rgb]{0.0,0,0}}

\usepackage{newfloat}
\usepackage{listings}

\DeclareCaptionStyle{ruled}{labelfont=normalfont,labelsep=colon,strut=off} 
\floatstyle{ruled}
\newfloat{listing}{tb}{lst}{}
\floatname{listing}{Listing}
\title{Hierarchical Cross-Modal Alignment for Open-Vocabulary 3D Object Detection}
\author {
    Youjun Zhao\equalcontrib,
    Jiaying Lin\equalcontrib,
    Rynson W.H. Lau\thanks{\ Corresponding author.}
}
\affiliations {
    Department of Computer Science, City University of Hong Kong\\ 
    youjun.zhao@my.cityu.edu.hk, jiayinlin5-c@my.cityu.edu.hk, Rynson.Lau@cityu.edu.hk
}
\usepackage{bibentry}

\begin{document}

\maketitle

\begin{abstract}
Open-vocabulary 3D object detection (OV-3DOD) aims at localizing and classifying novel objects beyond closed sets. The recent success of vision-language models (VLMs) has demonstrated their remarkable capabilities to understand open vocabularies. Existing works that leverage VLMs for 3D object detection (3DOD) generally resort to representations that lose the rich scene context required for 3D perception. To address this problem, we propose in this paper a hierarchical framework, named \textbf{HCMA}, to simultaneously learn local object and global scene information for OV-3DOD. Specifically, we first design a Hierarchical Data Integration (HDI) approach to obtain coarse-to-fine 3D-image-text data, which is fed into a VLM to extract object-centric knowledge. To facilitate the association of feature hierarchies, we then propose an Interactive Cross-Modal Alignment (ICMA) strategy to establish effective intra-level and inter-level feature connections. To better align features across different levels, we further propose an Object-Focusing Context Adjustment (OFCA) module to refine multi-level features by emphasizing object-related features. Extensive experiments demonstrate that the proposed method outperforms SOTA methods on the existing OV-3DOD benchmarks. It also achieves promising OV-3DOD results even without any 3D annotations.
\end{abstract}

\begin{links}
    \link{Code and Extended version}{https://youjunzhao.github.io/HCMA/}
\end{links}

\section{Introduction}
3D object detection (3DOD) aims at localizing and classifying objects in 3D scenes. As a fundamental task in 3D scene understanding, it has various potential applications, \textit{e.g.}, robotic~\cite{robotic} and autonomous driving~\cite{autodrive}. However, existing 3DOD methods are limited to a small set of categories due to the lack of annotated 3D datasets. 
Most 3DOD works focus on detecting seen categories that exist in the training data and cannot be easily generalized to real-world data with unseen categories. To address this limitation, open-vocabulary 3D object detection (OV-3DOD) aims to detect novel objects in 3D scenes that are outside the limited categories of the training dataset.

Recently, vision-language models (VLMs) such as CLIP \cite{clip} and ALIGN \cite{ALIGN} have made tremendous progress by training on large-scale image-text data from the Internet.
The ability of VLMs to learn the rich object-level context of image-text embedding has inspired open-vocabulary 2D understanding tasks, such as object detection \cite{vlid,HierKD,bangalath2022bridging,wang2023object} and semantic segmentation \cite{Zegformer,zhou2022extract,SAN,han2023global}. However, it is not straightforward to pre-train 3D-text models due to the lack of large-scale 3D-text data. Recent attempts \cite{ding2023pla,chen2023clip2scene,peng2023openscene} leverage rich linguistic representations of VLMs in 3D open-vocabulary understanding tasks by directly projecting 3D data onto the 2D image space. This motivates us to leverage VLMs for the OV-3DOD task. By doing so, we can utilize the rich object-centric context of linguistic descriptions in the 3D domain, allowing us to connect 3D point clouds with text representation for OV-3DOD.

However, directly applying 2D VLMs to address the 3DOD task is not feasible due to the significant contextual differences between 2D and 3D data. First, pre-trained 2D VLMs~\cite{chen2023protoclip} are typically trained on images containing mostly a single object. In contrast, 3D scenes often comprise multiple objects of various sizes. Connecting multiple objects with textual descriptions in a 3D scene is more challenging. Second, 3D scenes have more data dimensions than 2D images. This discrepancy may lead to the negligence of important spatial information associated with 3D objects. Third, unlike the 2D images used for pre-training VLMs, in which objects are usually located at the center, objects in a 3D scene may be located in various corners or even partially outside the 3D scene. This variability of 3D scenes makes it difficult to learn the model. The above three challenges pose a significant need for a holistic understanding of 3D scenes in the 3DOD task.

For effective 3D detection, our insight is that it is essential to consider not only the object features, but also the surrounding 3D scene context, in order to accurately locate the object. Hence, in this paper, we propose a hierarchical framework, named \textbf{HCMA}, to simultaneously learn 3D objects as well as scene contexts for the OV-3DOD task. Our HCMA framework has three technical contributions. 
First, unlike the state-of-the-art OV-3DOD method~\cite{ov-3det,cao2024coda}, which focuses solely on object-level contexts and aligns cross-modal features at the object level, we propose a novel Hierarchical Data Integration (HDI) approach to construct multi-level data, capturing hierarchical contexts in 3D scenes that incorporate object-level, view-level, and scene-level supervisions.
By introducing the global scene representation, the multi-level data enables a comprehensive understanding of 3D point cloud scenes.
Second, we design an Interactive Cross-Modal Alignment (ICMA) strategy to establish connections between the point cloud, image, and text representations in different hierarchies. ICMA contains two alignment approaches: intra-level cross-modal alignment and inter-level cross-modal alignment. The intra-level cross-modal alignment facilitates the integration of semantically related information within each hierarchy, while the inter-level cross-modal alignment compensates for the interaction between diverse hierarchies, enabling a coarse-to-fine understanding of the 3D scene. To further strengthen these connections across modalities, we employ a novel contrastive learning method to minimize the discrepancy among the representations derived from features in different modalities.
Third, we propose an Object-Focusing Context Adjustment (OFCA) module to achieve accurate and object-centric feature alignment across different hierarchies.
Instead of directly utilizing the features obtained from pre-trained VLMs~\cite{ov-3det,fmov3d,cao2024coda}, our OFCA takes an object-centric approach, leveraging object-related features and refining cross-level information to enable seamless context integration for objects across various levels and modalities. This helps alleviate the semantic bias inherited from VLMs and improves the robustness of the hierarchical features.

In summary, the main contributions of this paper include:
\begin{itemize}
    \item We propose \textbf{HCMA}, a hierarchical OV-3DOD framework to simultaneously learn object and 3D scene contexts to facilitate accurate 3D object detection. 
    \item We propose a Hierarchical Data Integration (\textbf{HDI}) approach to obtain coarse-to-fine cross-modal data for multi-level supervision. It provides a comprehensive understanding of contexts in 3D scenes.
    \item We design an effective Interactive Cross-Modal Alignment (\textbf{ICMA}) strategy to facilitate a coarse-to-fine feature interaction across different levels and modalities. To improve hierarchical feature alignments in ICMA, we further propose an Object-Focusing Context Adjustment (\textbf{OFCA}) module to facilitate the effective integration of contexts across different levels.
    \item We conduct extensive experiments on the OV-3DOD benchmarks. 
    Our method achieves superior performances compared to existing state-of-the-art approaches, demonstrating its effectiveness for the OV-3DOD task.
\end{itemize}

\section{Related Work}

\textbf{3D Object Detection.}
\rynn{This task} is to accurately locate all objects of interest in a given 3D scene. VoteNet \cite{votenet} first encodes point clouds into object volumetric representations and then connects to an RPN to generate detection results. Following \cite{votenet}, H3DNet \cite{h3dnet} integrates a set of geometric primitives to detect 3D objects. \jy{Recently,} the Transformer \cite{vaswani2017attention} is also \jy{found to be} suitable for 3D point cloud modeling since it is permutation-invariant and can capture long-range context. 3DETR \cite{3detr} proposes an end-to-end \rynn{Transformer model for 3D point cloud object detection}. \rynn{Unlike these \ryn{works}, which only utilize closed-set data to perform 3DOD, our work focuses on the open-vocabulary setting and proposes} a hierarchical framework to localize and classify novel 3D objects.

\vspace{1ex}

\noindent\textbf{Open-Vocabulary 3D Scene Understanding.}
Open-vocabulary 3D scene understanding includes tasks such as \jy{open-vocabulary} 3D segmentation and \jy{open-vocabulary 3D} object detection (OV-3DOD). \zyj{Compared with open-vocabulary 2D scene understanding~\cite{yoloworld,yang2024boosting}, research on open-vocabulary 3D scene understanding is limited}, \rynn{as 3D datasets are much more labor-intensive to create and therefore much smaller in scale}. PLA \cite{ding2023pla} associates 3D point clouds with text by generating text descriptions from captioning multi-view 3D scene images. It can perform open-vocabulary 3D segmentation in a 2D manner \jy{with the aligned 3D point cloud and text features}. CLIP2Scene \cite{chen2023clip2scene} utilizes the 2D vision-language model CLIP \cite{clip} for 3D scene understanding. It leverages \jy{cross-modal knowledge} from CLIP to extract semantic text features to achieve annotation-free 3D semantic segmentation. Similar to CLIP2Scene, OpenScene \cite{peng2023openscene} also infers CLIP features to obtain 3D-text co-embedding for open-vocabulary semantic segmentation.

Open-vocabulary 3D object detection (OV-3DOD) is another \rynn{task under} 3D scene understanding. OV-3DET~\cite{ov-3det} is one of the first works to extend the open-vocabulary object detection task into the 3D area. It leverages rich object context extracted from a \jy{2D} detector \jy{pre-trained on large-scale image datasets} and CLIP \cite{clip} to detect novel 3D objects. \zyjnew{CoDA \cite{cao2024coda} is a closer work to OV-3DET \cite{ov-3det}, but \jynew{it} does not rely on \rynn{an} additional 2D detector to help localize the 3D bounding box. Instead, it can simultaneously perform 3D object localization and classification. More recently, FM-OV3D \cite{fmov3d} tackles the OV-3DOD problem by blending knowledge from multiple pre-trained foundation models.} Our work is closer to \cite{ov-3det}, but we do not rely solely on \rynn{the} \jy{object-centered features} of point clouds for object detection. Our work can simultaneously learn hierarchical 3D scene \rynn{contexts} to align different modalities for discovering unseen 3D object patterns.
\begin{figure*}[tb] \centering

    \includegraphics[width=\textwidth]{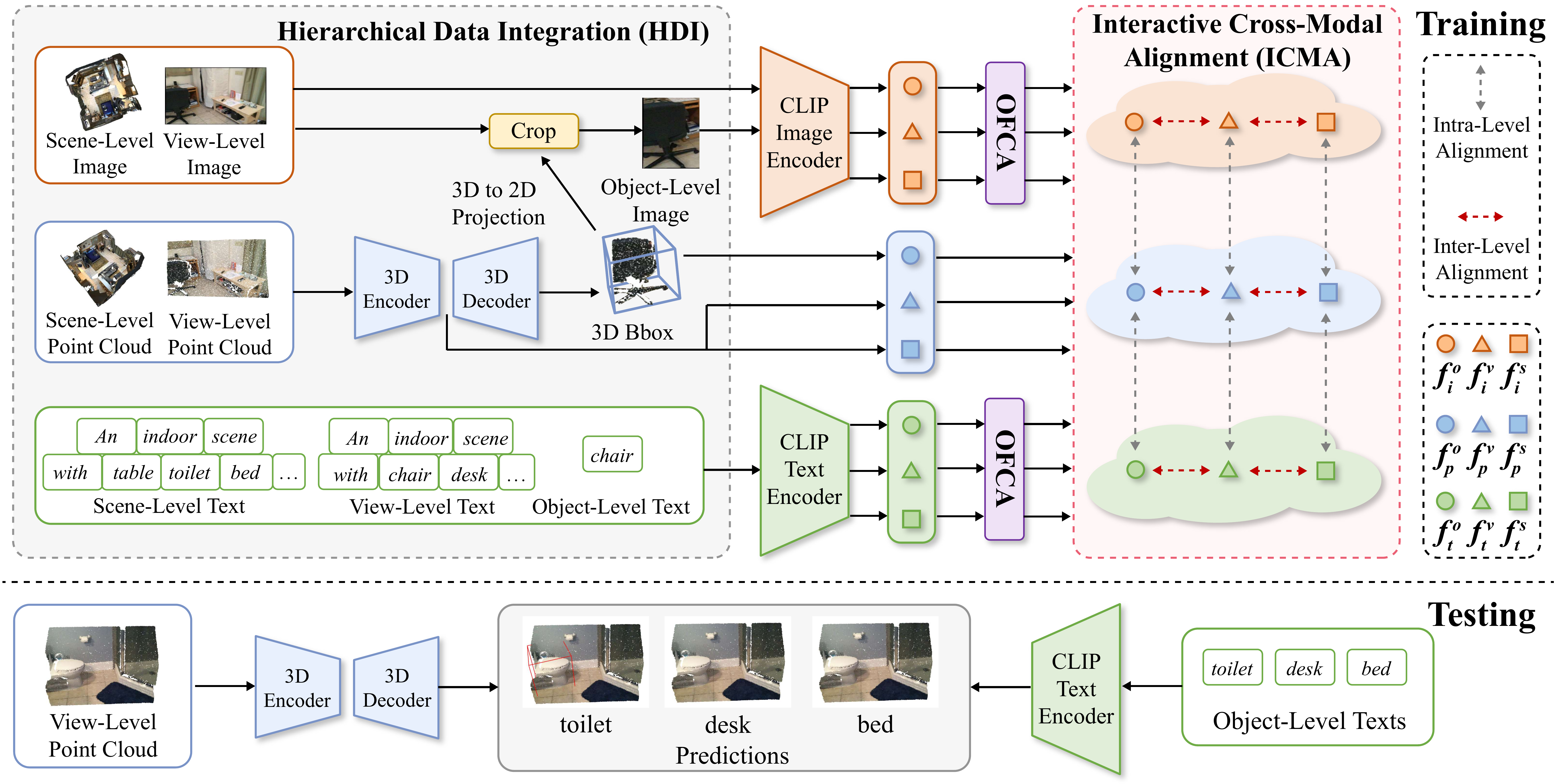}
    \vspace{-5mm}
    \caption{\rynn{Our HCMA framework is a three-stream network} containing \ryn{pre-trained CLIP image and text encoders}, a 3D detector, \zyjnew{and our proposed Object-Focusing Context Adjustment (OFCA) module.} Each stream processes three \rynn{types of} hierarchical semantics. \zyjnew{The input data of each stream is derived from our Hierarchical Data Integration (HDI) approach, while} the output semantics of each stream \rynn{is} associated \jy{by our} \zyjnew{Interactive Cross-Modal Alignment (ICMA) strategy.}
     }
    \label{framework_img}
    \vspace{-2mm}
\end{figure*}
\section{Method}

Our method, \ryn{named HCMA,} first learns to localize 3D objects from a pre-trained 2D object detector. \zyjnew{\rynn{To do this, we} generate 2D bounding boxes $B_{2D}$ of the view-level paired images $I^{v}$ from the pre-trained 2D detector. Prompted by $n$ training classes $C=\{C_{1}, C_{2}, ..., C_{n}\}$, the pre-trained 2D detector can generate 2D bounding boxes $B_{2D}\in\mathbb{R}^{4}$ of the training classes $C$ on the view-level paired images $I^{v}$ as:} 
\begin{equation}
{B}_{2D}=\mathrm{2DDetector}(I^{v}, C).
\label{eq1}
\end{equation}
Given predicted 2D bounding boxes $B_{2D}$, we \rynn{then} back-project \ryn{them} into the 3D space as 3D bounding boxes $B_{3D}\in\mathbb{R}^{7}$, \zyjnew{and \jynew{perform clustering to group points belonging to the same object}. 
\jynew{This step removes unrelated points and improves the accuracy of the resulting 3D bounding boxes.}
Subsequently, we generate 3D bounding boxes based on the remaining points as:}
\begin{equation}
B_{3D}=\mathrm{Cluster}(B_{2D} \cdot M^{-1}),
\label{eq2}
\end{equation}
\jynew{where $M$ is the projection matrix provided in the training dataset.}
We \jy{utilize} 3D bounding boxes $B_{3D}$ to supervise the localization of 3D objects \zyjnew{by computing a regression loss function between the predicted 3D bounding boxes $\hat{B}_{3D}$ and the target 3D bounding boxes $B_{3D}$ as:}
\begin{equation}
\mathcal{L}_{loc}=\mathcal{L}_{3D}(\hat{B}_{3D},B_{3D}).
\end{equation}
\rynn{HCMA next learns} to classify 3D localization results from text prompts with cross-modal contrastive learning. In \ryn{this way}, our method can perform 3D object detection without 3D annotations.
\zyjnew{\jynew{Figure~\ref{framework_img} shows the overall pipeline of our method.
During training, our method takes seven types of data as input, including three hierarchies and three modalities. \rynn{During inference, it takes only the raw view-level point clouds $P^{v}$ and object-level texts $T^{o}$ as input}.} Our method is a three-stream network consisting of a 3D point cloud object detector}
$\mathbf{H}_{p}$, 
an image encoder $\mathbf{H}_{i}$, a text encoder $\mathbf{H}_{t}$, \zyjnew{and our proposed OFCA module $\mathbf{H}_{o}$.} 
For each 3D-image-text tuple $(P,I,T)$ as input, \ryn{we extract the 3D-image-text features} $(f_{p},f_{i},f_{t})$ from the three-stream network as:
\begin{equation}
f_{p} = \mathbf{H}_{p}(P), \ \ \ f_{i} = \mathbf{H}_{o}(\mathbf{H}_{i}(I)), \ \ \ f_{t} = \mathbf{H}_{o}(\mathbf{H}_{t}(T)).
\end{equation}
Finally, we utilize hierarchical cross-modal contrastive learning to align $f_{p}$, $f_{i}$, and $f_{t}$. Note that we only update the \zyjnew{weights}
of the 3D detector and the OFCA module since we utilize \zyjnew{a pair of frozen CLIP image and text encoders}
in the training stage.

\subsection{\rynn{The HDI Approach}}

\begin{figure}[tb] 
\centering
    \includegraphics[width=0.47\textwidth]{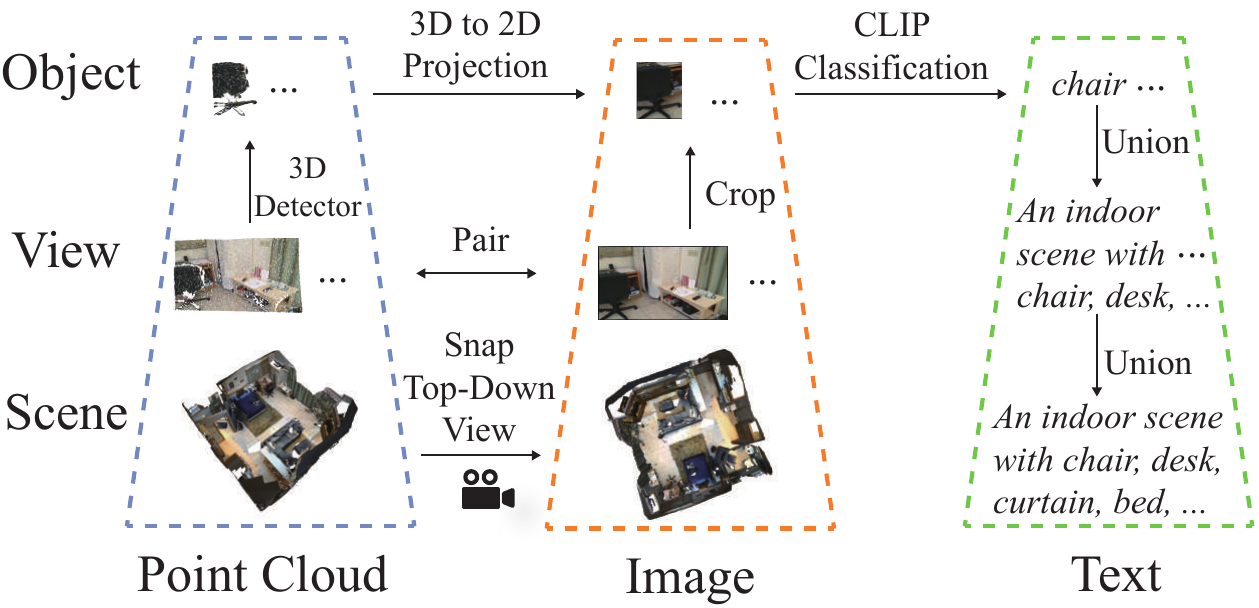}
    \vspace{-1mm}
    \caption{Illustration of \rynn{the \zyjnew{Hierarchical Data Integration (HDI)} approach}. HDI introduces object-level, view-level, and scene-level hierarchies to associate point \rynn{clouds, images, and texts}.} 
    \label{data_img}
    \vspace{-1mm}
\end{figure}

\ryn{We first} introduce the detailed process of \rynn{our \zyjnew{Hierarchical Data Integration (HDI)} approach}, as shown in Figure \ref{data_img}. Given a view-level 3D-image pair $(P^{v},I^{v})$, we can obtain object-level point cloud predictions from the 3D detector $\mathbf{H}_{p}$, including $m$ 3D bounding boxes and an object-level point cloud feature sequence, $f_{p}^{o}=(f_{p1}^{o}, f_{p2}^{o}, ..., f_{pm}^{o})$. 

We then project the predicted 3D bounding boxes into 2D bounding boxes with the camera calibration parameter. By cropping the view-level image $I^{v}$ with 2D bounding boxes, we can obtain \ryn{an object-level image set,} $I^{o}=(I^{o}_1, I^{o}_2, ..., I^{o}_{m})$, which is \ryn{then} sent into the \zyjnew{image encoder to get the object-level image features $f_i^{o}$. The object-level text set $T^{o}$ is sent into the text encoder to obtain the object-level text features $f_t^{o}$.}
These three features form the object-level 3D-image-text features $(f_p^{o},f_i^{o},f_t^{o})$.

For $m$ object-level text prompts, $T^{o}=(T^{o}_{1}, T^{o}_{2}, ..., T^{o}_{m})$, we concatenate them to form a view-level text caption $T^{v}$. \zyjnew{Since a 3D view point cloud $P^{v}$ contains $m$ 3D objects point cloud $P^{o}$, we provide a view-level text caption $T^{v}$ by merging and removing duplicates of these $m$ object-level texts.}
\jynew{Given a 3D view point cloud $P^v$ containing $m$ object point clouds $P^o$, we generate a view-level text caption $T^v$: }
\begin{equation}
T^{v}=(T^{o}_{1}\cup T^{o}_{2}\cup ...\cup T^{o}_{m}),
\end{equation}
where $\cup$ represents the set union operation. \jynew{It is worth noting that $T^{v}$ is constructed by consolidating the textual information from all object-level texts, ensuring the removal of any redundancies}. \rynn{It can also} provide a more fine-grained and accurate description of the view-level image $I^{v}$. \ryn{We then} feed the view-level 3D-image-text $(P^{v},I^{v}, T^{v})$ into their corresponding encoders and the OFCA module. In this way, we can generate the view-level 3D-image-text features $(f_p^{v},f_i^{v},f_t^{v})$.

In addition, we incorporate scene-level data to provide the coarse-grained context of the 3D scene. \zyj{Despite the absence of scene-level images in the 3D detection dataset, we address this limitation by generating a top-down image of the 3D scene as a scene-level image $I^{s}$ based on a scene-level 3D point cloud $P^{s}$.}
Since a 3D scene point cloud $P^{s}$ contains $n$ 3D views point cloud $P^{v}$, we provide a scene-level text caption $T^{s}$ by merging and removing duplicates of these $n$ view-level texts $T^v$. 
\jynew{Similarly, we have:}
\begin{equation}
T^{s}=(T^{v}_{1}\cup T^{v}_{2}\cup ...\cup T^{v}_{n}).
\end{equation}
Finally, we also feed the scene-level 3D-image-text $(P^{s},I^{s}, T^{s})$ into their corresponding encoders and the OFCA module to generate the scene-level 3D-image-text features $(f_p^{s},f_i^{s},f_t^{s})$.

It is worth mentioning that we only utilize point clouds and text in the inference stage. With object-level text $T^{o}$ as input, our method can \rynn{detect 3D objects} from the view-level point cloud $P^{v}$.
\subsection{\rynn{The ICMA Strategy}}
Our Interactive Cross-Modal Alignment (ICMA)  \rynn{strategy} consists of two parts: intra-level cross-modal alignment and inter-level cross-modal alignment.

\vspace{1ex}

\noindent\textbf{Intra-Level Cross-Modal Alignment.}
\ryn{Here}, we introduce the details of the intra-level cross-modal alignment in our proposed HCMA. As mentioned above, we have obtained multi-level features from object detector $\mathbf{H}_{p}$, text encoder $\mathbf{H}_{t}$, image encoder $\mathbf{H}_i$, and \zyjnew{OFCA module $\mathbf{H}_{o}$.} For the object-level features $(f_p^{o},f_i^{o},f_t^{o})$, we first classify them with \zyjnew{a category $\mathbb{C}^{o}$} from the text prompt with pre-trained CLIP:
\zyjnew{
\begin{equation}\label{textclass}
\mathbb{C}^{o} = \mathrm{argmax}(\mathrm{Softmax}(f_i^{o} \cdot f_t^{o})).
\end{equation}}
In this way, $(f_p^{o},f_t^{o})$ and $(f_p^{o},f_i^{o})$ with the same category $\mathbb{C}^{o}$ are used as two positive pairs. Conversely, they are considered two negative pairs if they \ryn{do not} match the category. In addition, there is no need to align $(f_i^{o},f_t^{o})$ again since they are already pre-aligned in the pre-trained CLIP model.

\begin{figure}[tb] \centering
    \includegraphics[width=0.46\textwidth]{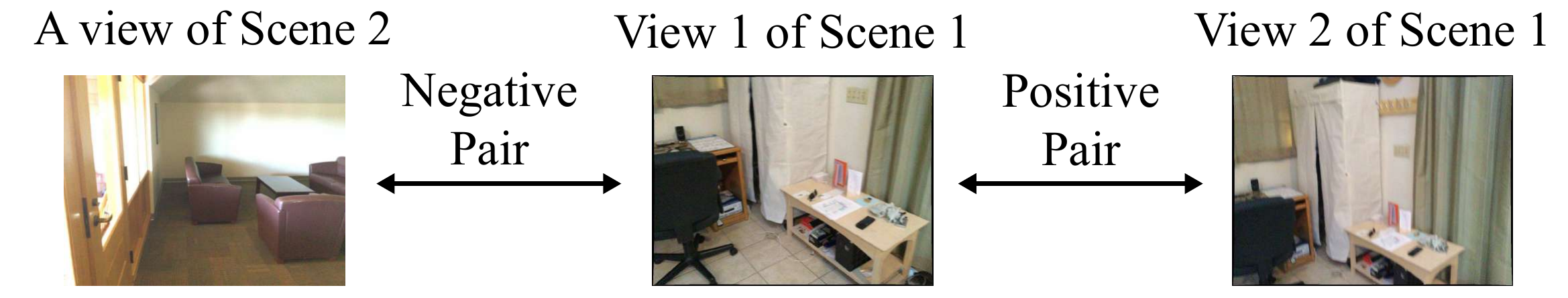}
    \vspace{-1mm}
    \caption{\rynn{Construction of the} positive and negative samples based on the scene label.} 
    \vspace{-1mm}
    \label{scene_label}
\end{figure}

For the view-level features $(f_p^{v},f_i^{v},f_t^{v})$, we classify \ryn{them} based on scene labels, as shown in Figure~\ref{scene_label}. Since an indoor scene can produce multiple view-level \ryn{point clouds and images} from different perspectives, it is natural that \ryn{these view-level point clouds and images from} the same scene are more similar to each other than to those from different scenes. Specifically, \ryn{if $(f_p^{v},f_t^{v})$ and $(f_p^{v},f_i^{v})$ have the same scene label, they form two} positive pairs. Otherwise, they are considered \rynn{as} two negative pairs.

Similarly, for the scene-level features $(f_p^{s},f_i^{s},f_t^{s})$, classification is also performed based on scene labels. We divide each of $(f_p^{s},f_t^{s})$ and $(f_p^{s},f_i^{s})$ into positive pairs and negative pairs according to their scene labels. 
This classification strategy based on scene labels allows for the grouping of similar view-level and scene-level features, enhancing the ability of our method to detect nearby objects in the same scene.

\vspace{1ex}

\noindent\textbf{Inter-Level Cross-Modal Alignment.}
To further improve the alignment precision and enhance the integration of information from multiple sources, we introduce \rynn{the inter-level cross-modal alignment here}. It enables coarse-to-fine feature interaction and fusion across different levels and modalities. Object-level input can provide the fine-grained context of the 3D object, while view-level and scene-level inputs are more coarse-grained. We propose three alignment strategies to capture the coarse-to-fine context of 3D objects, including local alignment, global alignment, and all alignment. These strategies aim to provide comprehensive contextual information to improve 3D object understanding.

For local alignment, we concatenate object-level features and view-level features into local features $(f_p^{l},f_i^{l},f_t^{l})$. \ryn{Local features contain both} object-level and view-level features. 
We then classify the local features with the text prompt category from the pre-trained CLIP, \zyjnew{similar to \rynn{Eq.}~\ref{textclass}}, \ryn{forming positive pairs and negative pairs from the 3D-text local features as well as from the 3D-image local features.}
Similar to local alignment, global alignment concatenates object-level features and scene-level features into global features $(f_p^{g},f_i^{g},f_t^{g})$. \ryn{Global features contain} object-level and scene-level features. 
\ryn{Again, we form positive pairs and negative pairs from the 3D-text global features as well as from the 3D-image global features according to the text category from the pre-trained CLIP, \zyjnew{similar to \rynn{Eq.}~\ref{textclass}.}}
In addition, we can also perform all alignment by concatenating object-level features, view-level features, and scene-level features into all features $(f_p^{a},f_i^{a},f_t^{a})$. All features contain object-level, view-level, and scene-level features.
\ryn{Finally, we form positive pairs and negative pairs from the 3D-text all features as well as from the 3D-image all features according to the text category from the pre-trained CLIP, similar to \rynn{Eq.}~\ref{textclass}.}

\subsection{\rynn{The OFCA Module}}

\begin{figure}[tb] \centering
    \includegraphics[width=0.44\textwidth]{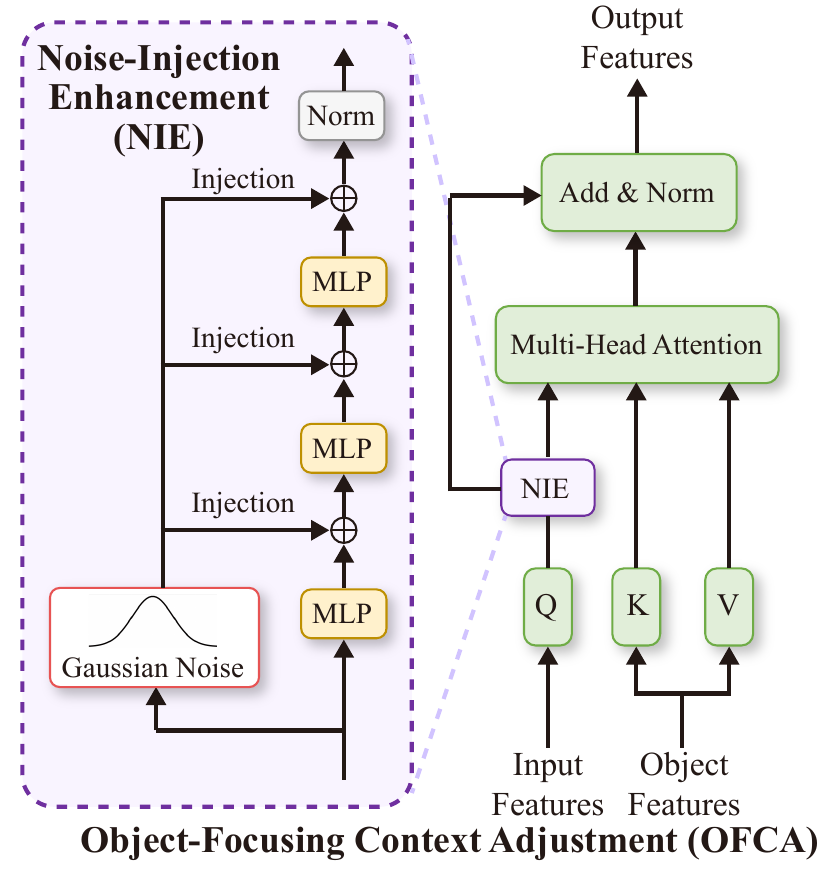}
    \caption{Structure of the proposed Object-Focusing Context Adjustment (OFCA) module.} 
    \label{ofca}
    \vspace{-1mm}
\end{figure}

\zyjnew{To better utilize the open-vocabulary capabilities of pre-trained VLMs in the OV-3DOD task, we introduce \rynn{the Object-Focusing Context Adjustment (OFCA) module, which helps refine} the features before cross-modal alignment. As illustrated in Figure~\ref{ofca}, 
\jy{each OFCA module consists of a Noise-Injection Enhancement (NIE) block and a multi-head attention layer. \zyj{The hierarchical semantic derived from object-centric VLMs inherit and amplify the object-level information during alignment in the ICMA module. The NIE block employs Gaussian noise to introduce perturbation to the semantic, which helps reduce potential semantic bias and improve robustness for information enhancement, ultimately improving the model's capability to generalize effectively to unseen data.} The NIE \rynn{block} refines input features and transforms them into the query for the attention layer, while the corresponding object features serve as the key and value. NIE employs Gaussian noise by gradually injecting it into the input features. Specifically, the NIE \rynn{block} is constructed by stacking $n$ identical MLP blocks.} With the input features $x_0$ and the output features $x_{m}$ from the $m^{th}$ MLP block, the $m^{th}$ Gaussian Noise injection operation can be written as:
\begin{equation}
x_{m}^{'} = \alpha(x_0 + \beta\frac{1}{\sqrt{2\pi}}e^{-\frac{x_0^2}{2}}) + (1-\alpha)x_m ,
\end{equation}
where $\alpha$ and $\beta$ control the weight of the injection operation. 
\jynew{Note that we only input view-level and scene-level features into the NIE block and strategically exclude object-level features from the NIE \rynn{block, as} object-level features are often more reliable compared to other-level features from the pre-trained VLMs.}}

\subsection{Hierarchical Cross-Modal Contrastive Learning}

Using the obtained 3D-text features $(f_p,f_t)$ and 3D-image features $(f_p,f_i)$, we can guide the 3D detector $\mathbf{H}_{p}$ to detect novel objects by learning from coarse-to-fine supervisions. We introduce a general contrastive learning method to achieve consistency between 3D, visual, and linguistic representations of hierarchical semantics. The main contrastive loss function \cite{oord2018representation} is given by:
\begin{equation}
\mathcal{L}_{c}=-\frac{1}{B}\sum_{b=1}^{B}log\frac{\begin{matrix} \sum_{i=0}^m e^{s_{b}^{T}s_{i}/\tau}\end{matrix}}{\begin{matrix} \sum_{j=0}^B e^{s_{b}^{T}s_{j}/\tau}\end{matrix}},
\end{equation}
where $\tau$ is a temperature parameter, $s$ denotes the samples in contrastive learning, $B$ is the number of samples in each batch, and $m$ is the number of positive samples.

For 3D-text pairs $(f_p,f_t)$ and 3D-image pairs $(f_p,f_i)$, \zyjnew{we can compute the contrastive loss separately.} The cross-modal contrastive loss is given by:
\begin{equation}
\mathcal{L}_{m}=\mathcal{L}_{c}(f_p,f_t)+\mathcal{L}_{c}(f_p,f_i).
\end{equation}
Subsequently, we employ the positive and negative samples mentioned earlier to construct six coarse-to-fine supervisions. These include \zyj{object-level $\mathcal{L}^{o}_{m}$, view-level $\mathcal{L}^{v}_{m}$, and scene-level $\mathcal{L}^{s}_{m}$} derived from inter-level cross-modal alignment. Additionally, we incorporate \zyj{local $\mathcal{L}^{l}_{m}$, global $\mathcal{L}^{g}_{m}$, and all $\mathcal{L}^{a}_{m}$} supervisions through inter-level cross-modal alignment. The alignment loss is given by:
\zyj{
\begin{equation}
\mathcal{L}_{align}= \mathds{1} \cdot \left[\mathcal{L}_{m}^{o},\mathcal{L}_{m}^{v},\mathcal{L}_{m}^{s},\mathcal{L}_{m}^{l},\mathcal{L}_{m}^{g},\mathcal{L}_{m}^{a}\right],
\label{loss_control}
\end{equation}
}
\zyj{where $\mathds{1}$ is the indicator function} that controls the specific form of $\mathcal{L}_{align}$, which is decided by different hierarchies utilized in the training stage. The overall training objective can be written as:
\begin{equation}
\mathcal{L}=\mathcal{L}_{align}+ \mathcal{L}_{loc},
\end{equation}
where $\mathcal{L}_{loc}$ is also used to supervise the 3D detector.

\begin{table*}[tb]\centering
    \resizebox{\textwidth}{!}{
    \large
    \begin{tabular}{c|c|cccccccccccccccccccc}
    \toprule
        Method & $mAP_{25}$ & \rotatebox{90}{toilet} & \rotatebox{90}{bed} & \rotatebox{90}{chair} & \rotatebox{90}{sofa} & \rotatebox{90}{dresser} & \rotatebox{90}{table} & \rotatebox{90}{cabinet} & \rotatebox{90}{bookshelf} & \rotatebox{90}{pillow} & \rotatebox{90}{sink} & \rotatebox{90}{bathtub} & \rotatebox{90}{refrigerator} & \rotatebox{90}{desk} & \rotatebox{90}{night stand} & \rotatebox{90}{counter} & \rotatebox{90}{door} & \rotatebox{90}{curtain} & \rotatebox{90}{box} & \rotatebox{90}{lamp} & \rotatebox{90}{bag}\\
    \midrule
   OV-PointCLIP & 0.74 & 1.04 & 1.85 & 4.79 & 1.18 & 0.19 & 1.61 & 0.41 & 0.03 & 0.40 & 0.29 & 0.51 & 0.10 & 1.66 & 0.16 & 0.02 & 0.24 & 0.04 & 0.15 & 0.03 & 0.05 \\
         OV-3DETIC  & 14.99 & 53.26 & 24.88 & 15.77 & 31.36 & 11.54 & 9.14 & 2.10 & 9.39 & 17.00 & 29.21 & 27.45 & 19.96 & 13.68 & 0.01 & 0.00 & 0.00 & 17.73 & 4.80 & 3.04 & 9.51 \\
        OV-CLIP-3D & 12.68 & 44.78 & 23.84 & 17.52 & 12.62 & 4.92 & 13.24 & 1.95 & 3.97 & 11.37 & 17.64 & 32.24 & 14.87 & 11.38 & 2.37 & 0.51 & 14.46 & 8.58 & 7.45 & 5.14 & 4.70 \\
        OV-3DET & 18.02 & 57.29 & 42.26 & 27.06 & 31.50 & 8.21 & 14.17 & 2.98 & 5.56 & 23.00 & 31.60 & 56.28 & 10.99 & 19.72 & 0.77 & 0.31 & 9.59 & 10.53 & 3.78 & 2.11 & 2.71 \\
        CoDA & 19.32 & 68.09 & 44.04 & 28.72 & 44.57 & 3.41 & 20.23 & 5.32 & 0.03 & 27.95 & 45.26 & 50.51 & 6.55 & 12.42 & 15.15 & 0.68 & 7.95 & 0.01 & 2.94 & 0.51 & 2.02 \\
        \midrule
        \textbf{HCMA} (Ours) & \textbf{21.77} & 72.85 & 50.61 & 37.26 & 56.82 & 3.11 & 15.19 & 3.10 & 11.78 & 20.89 & 44.51 & 50.13 & 9.49 & 12.73 & 24.59 & 0.10 & 13.56 & 0.19 & 3.27 & 3.28 & 1.86 \\
        \bottomrule
    \end{tabular}
    }
    \vspace{-1mm}
    
    \caption{\zyjnew{Results on ScanNet in terms of $AP_{25}$. We report \jy{the average} value of all 20 categories.}}
    \vspace{-1mm}
    
    \label{main_scannet}
\end{table*}

\begin{table*}[tb]\centering
    \resizebox{\textwidth}{!}{
    \large
    \begin{tabular}{c|c|cccccccccccccccccccc}
    \toprule
        Method & $mAP_{25}$ & \rotatebox{90}{toilet} &\rotatebox{90}{bed} & \rotatebox{90}{chair} & \rotatebox{90}{bathtub} & \rotatebox{90}{sofa} & \rotatebox{90}{dresser} & \rotatebox{90}{scanner} & \rotatebox{90}{fridge} & \rotatebox{90}{lamp} & \rotatebox{90}{desk} & \rotatebox{90}{table} & \rotatebox{90}{stand} & \rotatebox{90}{cabinet} & \rotatebox{90}{counter} & \rotatebox{90}{bin} & \rotatebox{90}{bookshelf} & \rotatebox{90}{pillow} & \rotatebox{90}{microwave} & \rotatebox{90}{sink} & \rotatebox{90}{stool}\\
    \midrule
    OV-PointCLIP & 1.04 & 4.76 & 0.91 & 4.41 & 0.07 & 4.11 & 0.15 & 0.05 & 0.19 & 0.08 & 1.11 & 2.13 & 0.10 & 0.24 & 0.02 & 0.96 & 0.06 & 0.43 & 0.03 & 0.91 & 0.05 \\
    OV-3DETIC & 12.99 &  47.68 &41.35 &4.46& 24.41& 18.58 &10.42 &3.72 &5.74 &12.60 &4.89 &1.54 &0.00 &1.24 &0.00 &19.33& 4.58& 12.30& 23.78 &22.02 &1.17 \\
   OV-CLIP-3D & 11.80&  38.05& 34.45& 16.26& 20.07& 12.72& 8.03& 2.61 &14.62& 10.02& 5.26 &12.31& 4.02& 1.26& 0.09& 26.50& 7.78& 6.52& 4.28 &10.00& 1.19\\
   OV-3DET & 20.46 & 72.64 &66.13 &34.80& 44.74 &42.10 &11.52 &0.29 &12.57 &14.64 &11.21& 23.31 &2.75 &3.40& 0.75& 23.52& 9.83& 10.27& 1.98 &18.57& 4.10 \\
    \midrule
    \textbf{HCMA} (ours) &\textbf{21.53} & 68.50 & 72.81 & 40.59 & 49.65 & 43.20 & 3.32 & 0.03 & 17.38 & 14.34 & 11.73 & 28.61 & 19.48 & 0.56 & 0.12 & 11.33 & 1.51 & 10.34 & 1.34 & 31.56 & 4.10 \\
   \bottomrule
    \end{tabular}
    }
    \vspace{-1mm}
    
    \caption{\zyjnew{Results on SUN RGB-D in terms of $AP_{25}$. We report \jy{the average} value of all 20 categories.}}
    \vspace{-1mm}
    
    \label{main_sunrgbd}
\end{table*}

\section{Experiments}
\subsection{Datasets and Metrics} 
ScanNet \cite{dai2017scannet} is a widely used 3D object detection dataset. As our method requires the view-level and scene-level 3D-image pairs as input, we utilize the raw view-level point clouds, view-level image, and scene-level point clouds from ScanNet. We then generate the scene-level images by snapping the top-down view of the scene-level point clouds. \zyjnew{SUN RGB-D~\cite{song2015sunrgbd} is another popular 3D object detection dataset. Since SUN RGB-D dataset lacks scene-level point cloud data, we leverage the available raw view-level point clouds and view-level images from SUN RGB-D for 3D detection.} We have conducted our experiment on these two datasets to validate the effectiveness of our method in 20 vocabularies.
As for the evaluation metrics, we use average precision (AP) and mean average precision (mAP) at IoU thresholds of 0.25 and 0.5, denoted as $AP_{25}$, $mAP_{25}$, and $mAP_{50}$, respectively.

\subsection{Implementation Details}
\zyjnew{Our experimental setup} follows that of OV-3DET \cite{ov-3det} for fair comparison. We train our model \rynn{using} the AdamW optimizer with a cosine learning rate scheme. The base learning rate and the weight decay are set to $10^{-4}$ and $0.1$, respectively. The temperature parameter $\tau$ is set to 0.1 in contrastive learning. We adopt 3DETR \cite{3detr} as our 3D detector backbone. The number of object queries for 3DETR is set to 128. Experiments are conducted on a single RTX4090 GPU. \zyjnew{Our training epoch is the same as the baseline method OV-3DET.}
\subsection{Main Results}
\zyjnew{We evaluate the performance of our method by comparing it with previous approaches on both ScanNet and SUN RGB-D benchmarks. We conduct evaluations on 20 common classes to assess the effectiveness of our method.} Since open-vocabulary 3D object detection is still a new task, very few works can be \rynn{used for direct comparison}. We first directly compare our method with the reported results by OV-3DET \cite{ov-3det}, which is a strong baseline method for OV-3DOD. Following OV-3DET \cite{ov-3det}, we evaluate our method by adapting some existing works on open-vocabulary 2D object detection and 3D object classification into our setting. The baseline methods include PointCLIP \cite{zhang2022pointclip}, 3DETIC \cite{3detic}, and CLIP-3D \cite{clip} to conduct open-vocabulary 3D object detection, denoted as OV-PointCLIP \cite{zhang2022pointclip}, OV-3DETIC \cite{3detic}, and OV-CLIP-3D \cite{clip}. We also compare our method with the reported results by CoDA~\cite{cao2024coda}, which is the state-of-the-art method for OV-3DOD. Results are presented in Table~\ref{main_scannet} and Table~\ref{main_sunrgbd}. Our method shows superior \rynn{performances}, outperforming the previous methods \rynn{on both ScanNet and SUN RGB-D datasets in the OV-3DOD task}. 

\begin{figure*}[tb] \centering
    \includegraphics[width=.9\textwidth]{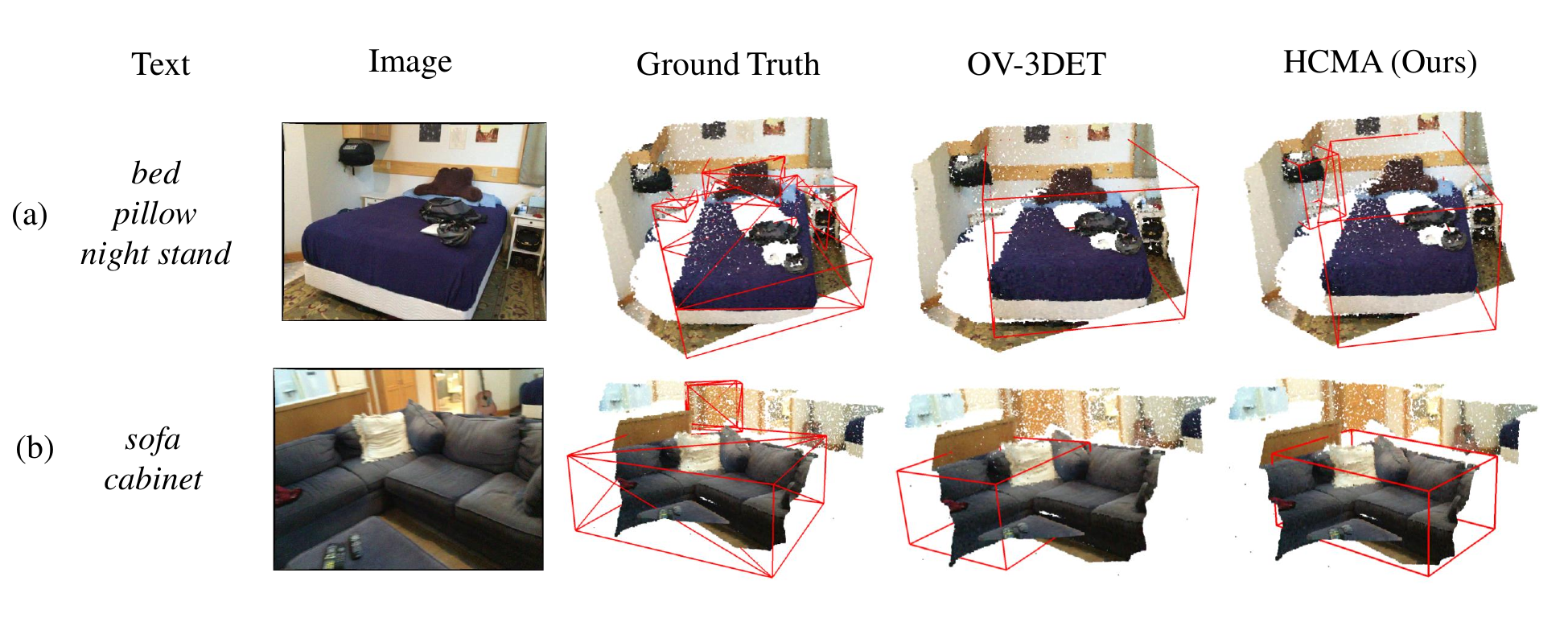}
    \vspace{-4mm}
    \caption{Qualitative comparison with OV-3DET. Our \rynn{HCMA framework} can perform more accurate OV-3DOD and can detect a wider vocabulary of objects in 3D scenes. For each case, the detection text prompts are shown on the left.} 
    \label{visual_img}
    \vspace{-1mm}
\end{figure*}

\noindent\textbf{Qualitative Results.}
\rynn{Figure \ref{visual_img} shows the qualitative results of} our method. Our HCMA \rynn{framework} can generate more accurate bounding boxes compared with the baseline OV-3DET \cite{ov-3det}.
HCMA can predict the location, size, and orientation of 3D bounding boxes more precisely. Specific instances, such as the sofa in sample $b$, highlight the superior performance of HCMA. This is because HCMA can provide coarse-gained context around the 3D object, allowing it to perceive a more complete object structure. 
Another advantage of our HCMA \rynn{framework}, as compared with the baseline method OV-3DET, is the ability to detect a wider range of objects with diverse vocabularies in 3D scenes. For example, the nightstand in sample $a$ is an \zyj{occluded} object missed by the baseline method. The hierarchical structure of our HCMA facilitates the understanding of objects in the same 3D scene by learning the contextual information, thus enhancing the overall object detection performance.

\subsection{Ablation Study}

\zyj{
\textbf{Overall Analysis.}
We conduct an ablation experiment on the ScanNet dataset to analyze the effectiveness of our proposed components. Starting with the base method that leverages solely object-level features and intra-level cross-modal alignment in our framework, we gradually incorporate our proposed modules. The results are summarized in Table~\ref{ablation_all} and demonstrate that each component contributes to the final result. By employing the HDI approach with both object-level and view-level feature\jynn{s}, the ICMA strategy with both intra-level and inter-level alignment, and the OFCA module with the NIE block, we observe improvements respectively in terms of $mAP_{25}$, and $mAP_{50}$.
}

\noindent\textbf{Analysis of the HDI Approach.}
We discuss the impact of the HDI approach through ablation studies on ScanNet. Specifically, we have three coarse-to-fine hierarchical supervisions during the training stage, including \zyj{object (O), view (V), and scene (S) levels.} Since the target of 3D object detection is to detect object-level point clouds in a 3D scene, object-level supervision is naturally the basis of object detection. Hence, we preserve object-level hierarchy to explore the impact of view-level and scene-level hierarchies in the ablation study. We employ four training strategies in our framework and analyze the results in Table~\ref{ablation_hdi}. \zyj{The results demonstrate that view-level (V), and scene-level (S) hierarchies can bring significant improvement individually.} The coarse-gained cross-modal correspondence is crucial for 3D detection. By leveraging the hierarchical structure and incorporating cross-modal information, our framework benefits from a more comprehensive understanding of the \zyj{3D scenes}.

\vspace{1ex}

\noindent\textbf{\zyjnew{Analysis of \rynn{the ICMA Strategy}.}}
We investigate the effectiveness of our ICMA strategy through an ablation study on ScanNet, which includes intra-level and inter-level cross-modal alignment. \rynn{The results are} shown in Table \ref{ablation_icma}. Specifically, we compare our method with three strategies: intra-level, inter-level, and both. The experiments are conducted on both object-level and view-level hierarchies. Results show that using both intra-level and inter-level cross-modal alignments outperforms the other two strategies in terms of $mAP_{25}$ and $mAP_{50}$, which indicates that both intra-level and inter-level alignment can contribute to cross-modal connections. The intra-level alignment provides hierarchical information to the inter-level alignment, while the inter-level interaction can compensate for the lack of coarse-to-fine context in the intra-level alignment. This complementary relationship enhances the overall performance by incorporating both local and global contextual information. 

\vspace{1ex}

\noindent\textbf{Analysis of \rynn{the OFCA} Module.}
\jynew{To assess the effectiveness of the OFCA module, we \rynn{conduct} ablation studies on the ScanNet dataset as shown in Table~\ref{ablation_ofca}. Experiments are conducted on both object-level and view-level hierarchies. The results show that the performance improves when employing the OFCA module, as evidenced by higher $mAP_{25}$ and $mAP_{50}$ compared to the baseline without the OFCA module (``w/o OFCA''). This suggests that the proposed OFCA effectively learns more accurate features by adjusting the cross-level features. We further analyze the impact of the NIE block within the OFCA module. Including the NIE block leads to further performance gains compared to using OFCA alone (``OFCA w/o NIE''). This indicates that the NIE block in the OFCA module contributes to performance improvement by enhancing the robustness at the feature level.}

\begin{table}[t]
\centering
\renewcommand\arraystretch{1.1}

\begin{minipage}{0.26\textwidth}
    \centering
    \resizebox{\textwidth}{!}{
        \setlength{\tabcolsep}{1.1mm}
        {
        \begin{tabular}{ccccc}
            \toprule
            HDI & ICMA & OFCA & $mAP_{25}$ & $mAP_{50}$ \\
            \midrule
            & & & 19.37 & 5.18 \\
            \checkmark & & & 19.71 & 5.80 \\
            \checkmark & \checkmark & & 20.18 & 6.11 \\
            \checkmark & \checkmark & \checkmark & \textbf{20.84} & \textbf{6.53} \\
            \bottomrule
        \end{tabular}
    }
    }
    \caption{\zyj{Ablation study of our designs.}}
    \label{ablation_all}
\end{minipage}
    \vspace{-1mm}
\hfill
\begin{minipage}{0.2\textwidth}
    \centering
    \resizebox{\textwidth}{!}{
            \setlength{\tabcolsep}{1.1mm}
            {
        \begin{tabular}{ccccc}
            \toprule
            O & V & S & $mAP_{25}$ & $mAP_{50}$ \\
            \midrule
            \checkmark & & & 19.37 & 5.18 \\
            \checkmark & \checkmark & & 20.18 & 6.11 \\
            \checkmark & & \checkmark & 20.20 & 6.18 \\
            \checkmark & \checkmark & \checkmark & \textbf{20.30} & \textbf{6.39} \\
            \bottomrule
        \end{tabular}
    }
    }
    \caption{Ablation study of the HDI approach.}
    \label{ablation_hdi}
\end{minipage}

\end{table}

\begin{table}[t]
\centering
\renewcommand\arraystretch{1.4}

\begin{minipage}{0.22\textwidth}
    \centering
    \resizebox{\textwidth}{!}{
        \begin{tabular}{cccc}
            \toprule
            Intra & Inter & $mAP_{25}$ & $mAP_{50}$ \\
            \midrule
            \checkmark &  & 19.71 & 5.80  \\
            & \checkmark  & 19.68 & 5.82  \\
            \checkmark & \checkmark  & \textbf{20.18} & \textbf{6.11}  \\
            \bottomrule
        \end{tabular}
    }
    \caption{Ablation study of the ICMA strategy.}
    \label{ablation_icma}
\end{minipage}
\hfill
\begin{minipage}{0.24\textwidth}
    \centering
    \resizebox{\textwidth}{!}{
        \begin{tabular}{lcc}
            \toprule
            Method & $mAP_{25}$ & $mAP_{50}$ \\
            \midrule
            w/o OFCA & 20.18 & 6.11 \\
            OFCA w/o NIE & 20.55 & 6.33 \\
            OFCA & \textbf{20.84} & \textbf{6.53} \\
            \bottomrule
        \end{tabular}
    }
    \caption{Ablation study of the OFCA module.}
    \label{ablation_ofca}
\end{minipage}
    \vspace{-2mm}
\end{table}

\section{Conclusion}

In this paper, we have introduced HCMA, a hierarchical framework to address fundamental \rynn{detection issues of} OV-3DOD. HCMA is designed to enhance the accuracy of cross-modal alignment of 3D, image, and text modalities. \zyjnew{We first introduce \rynn{the Hierarchical Data Integration (HDI) approach} to obtain hierarchical context from object, view, and scene levels.} \zyjnew{This coarse-to-fine supervision} enables HCMA to obtain more accurate results. \rynn{In addition}, \zyjnew{we design an effective Interactive Cross-Modal Alignment (ICMA) strategy, along with the Object-Focusing Context Adjustment (OFCA) module,} to align features of 3D, images, and texts to achieve more robust performances. Extensive experimental results demonstrate the superiority of HCMA in OV-3DOD.

\bibliography{aaai25}

\begin{thebibliography}{33}
\providecommand{\natexlab}[1]{#1}

\bibitem[{Bangalath et~al.(2022)Bangalath, Maaz, Khattak, Khan, and Shahbaz~Khan}]{bangalath2022bridging}
Bangalath, H.; Maaz, M.; Khattak, M.~U.; Khan, S.; and Shahbaz~Khan, F. 2022.
\newblock Bridging the gap between object and image-level representations for open-vocabulary detection.
\newblock In \emph{Proceedings of the Advances in Neural Information Processing Systems}, 33781--33794.

\bibitem[{Bojarski et~al.(2016)Bojarski, Del~Testa, Dworakowski, Firner, Flepp, Goyal, Jackel, Monfort, Muller, Zhang et~al.}]{autodrive}
Bojarski, M.; Del~Testa, D.; Dworakowski, D.; Firner, B.; Flepp, B.; Goyal, P.; Jackel, L.~D.; Monfort, M.; Muller, U.; Zhang, J.; et~al. 2016.
\newblock End to end learning for self-driving cars.
\newblock \emph{arXiv:1604.07316}.

\bibitem[{Cao et~al.(2024)Cao, Yihan, Xu, and Xu}]{cao2024coda}
Cao, Y.; Yihan, Z.; Xu, H.; and Xu, D. 2024.
\newblock Coda: Collaborative novel box discovery and cross-modal alignment for open-vocabulary 3d object detection.
\newblock In \emph{Proceedings of the Advances in Neural Information Processing Systems}, volume~36.

\bibitem[{Chen et~al.(2023{\natexlab{a}})Chen, Wu, Liu, Yang, Zheng, Tan, and Zhou}]{chen2023protoclip}
Chen, D.; Wu, Z.; Liu, F.; Yang, Z.; Zheng, S.; Tan, Y.; and Zhou, E. 2023{\natexlab{a}}.
\newblock ProtoCLIP: Prototypical Contrastive Language Image Pretraining.
\newblock \emph{IEEE Transactions on Neural Networks and Learning Systems}.

\bibitem[{Chen et~al.(2023{\natexlab{b}})Chen, Liu, Kong, Zhu, Ma, Li, Hou, Qiao, and Wang}]{chen2023clip2scene}
Chen, R.; Liu, Y.; Kong, L.; Zhu, X.; Ma, Y.; Li, Y.; Hou, Y.; Qiao, Y.; and Wang, W. 2023{\natexlab{b}}.
\newblock CLIP2Scene: Towards Label-efficient 3D Scene Understanding by CLIP.
\newblock In \emph{Proceedings of the IEEE/CVF Conference on Computer Vision and Pattern Recognition}, 7020--7030.

\bibitem[{Cheng et~al.(2024)Cheng, Song, Ge, Liu, Wang, and Shan}]{yoloworld}
Cheng, T.; Song, L.; Ge, Y.; Liu, W.; Wang, X.; and Shan, Y. 2024.
\newblock Yolo-world: Real-time open-vocabulary object detection.
\newblock In \emph{Proceedings of the IEEE/CVF Conference on Computer Vision and Pattern Recognition}, 16901--16911.

\bibitem[{Dai et~al.(2017)Dai, Chang, Savva, Halber, Funkhouser, and Nie{\ss}ner}]{dai2017scannet}
Dai, A.; Chang, A.~X.; Savva, M.; Halber, M.; Funkhouser, T.; and Nie{\ss}ner, M. 2017.
\newblock Scannet: Richly-annotated 3d reconstructions of indoor scenes.
\newblock In \emph{Proceedings of the IEEE Conference on Computer Vision and Pattern Recognition}, 5828--5839.

\bibitem[{Ding et~al.(2022)Ding, Xue, Xia, and Dai}]{Zegformer}
Ding, J.; Xue, N.; Xia, G.-S.; and Dai, D. 2022.
\newblock Decoupling zero-shot semantic segmentation.
\newblock In \emph{Proceedings of the IEEE/CVF Conference on Computer Vision and Pattern Recognition}, 11583--11592.

\bibitem[{Ding et~al.(2023)Ding, Yang, Xue, Zhang, Bai, and Qi}]{ding2023pla}
Ding, R.; Yang, J.; Xue, C.; Zhang, W.; Bai, S.; and Qi, X. 2023.
\newblock PLA: Language-Driven Open-Vocabulary 3D Scene Understanding.
\newblock In \emph{Proceedings of the IEEE/CVF Conference on Computer Vision and Pattern Recognition}, 7010--7019.

\bibitem[{Gu et~al.(2022)Gu, Lin, Kuo, and Cui}]{vlid}
Gu, X.; Lin, T.-Y.; Kuo, W.; and Cui, Y. 2022.
\newblock Open-vocabulary Object Detection via Vision and Language Knowledge Distillation.
\newblock In \emph{Proceedings of the International Conference on Learning Representations}.

\bibitem[{Gupta, Dollar, and Girshick(2019)}]{gupta2019lvis}
Gupta, A.; Dollar, P.; and Girshick, R. 2019.
\newblock Lvis: A dataset for large vocabulary instance segmentation.
\newblock In \emph{Proceedings of the IEEE/CVF Conference on Computer Vision and Pattern Recognition}, 5356--5364.

\bibitem[{Han et~al.(2023)Han, Liu, Liew, Ding, Liu, Wang, Tang, Yang, Feng, Zhao et~al.}]{han2023global}
Han, K.; Liu, Y.; Liew, J.~H.; Ding, H.; Liu, J.; Wang, Y.; Tang, Y.; Yang, Y.; Feng, J.; Zhao, Y.; et~al. 2023.
\newblock Global knowledge calibration for fast open-vocabulary segmentation.
\newblock In \emph{Proceedings of the IEEE/CVF International Conference on Computer Vision}, 797--807.

\bibitem[{Jia et~al.(2021)Jia, Yang, Xia, Chen, Parekh, Pham, Le, Sung, Li, and Duerig}]{ALIGN}
Jia, C.; Yang, Y.; Xia, Y.; Chen, Y.-T.; Parekh, Z.; Pham, H.; Le, Q.; Sung, Y.-H.; Li, Z.; and Duerig, T. 2021.
\newblock Scaling up visual and vision-language representation learning with noisy text supervision.
\newblock In \emph{Proceedings of the International conference on Machine Learning}, 4904--4916.

\bibitem[{Lu et~al.(2022)Lu, Xu, Wei, Xie, Tomizuka, Keutzer, and Zhang}]{ov-3detic}
Lu, Y.; Xu, C.; Wei, X.; Xie, X.; Tomizuka, M.; Keutzer, K.; and Zhang, S. 2022.
\newblock Open-vocabulary 3d detection via image-level class and debiased cross-modal contrastive learning.
\newblock \emph{arXiv:2207.01987}.

\bibitem[{Lu et~al.(2023)Lu, Xu, Wei, Xie, Tomizuka, Keutzer, and Zhang}]{ov-3det}
Lu, Y.; Xu, C.; Wei, X.; Xie, X.; Tomizuka, M.; Keutzer, K.; and Zhang, S. 2023.
\newblock Open-vocabulary point-cloud object detection without 3d annotation.
\newblock In \emph{Proceedings of the IEEE/CVF Conference on Computer Vision and Pattern Recognition}, 1190--1199.

\bibitem[{Ma et~al.(2022)Ma, Luo, Gao, Li, Chen, Wang, Zhang, and Hu}]{HierKD}
Ma, Z.; Luo, G.; Gao, J.; Li, L.; Chen, Y.; Wang, S.; Zhang, C.; and Hu, W. 2022.
\newblock Open-vocabulary one-stage detection with hierarchical visual-language knowledge distillation.
\newblock In \emph{Proceedings of the IEEE/CVF Conference on Computer Vision and Pattern Recognition}, 14074--14083.

\bibitem[{Misra, Girdhar, and Joulin(2021)}]{3detr}
Misra, I.; Girdhar, R.; and Joulin, A. 2021.
\newblock An end-to-end transformer model for 3d object detection.
\newblock In \emph{Proceedings of the IEEE/CVF International Conference on Computer Vision}, 2906--2917.

\bibitem[{Oord, Li, and Vinyals(2018)}]{oord2018representation}
Oord, A. v.~d.; Li, Y.; and Vinyals, O. 2018.
\newblock Representation learning with contrastive predictive coding.
\newblock \emph{arXiv:1807.03748}.

\bibitem[{Peng et~al.(2023)Peng, Genova, Jiang, Tagliasacchi, Pollefeys, Funkhouser et~al.}]{peng2023openscene}
Peng, S.; Genova, K.; Jiang, C.; Tagliasacchi, A.; Pollefeys, M.; Funkhouser, T.; et~al. 2023.
\newblock Openscene: 3d scene understanding with open vocabularies.
\newblock In \emph{Proceedings of the IEEE/CVF Conference on Computer Vision and Pattern Recognition}, 815--824.

\bibitem[{Qi et~al.(2019)Qi, Litany, He, and Guibas}]{votenet}
Qi, C.~R.; Litany, O.; He, K.; and Guibas, L.~J. 2019.
\newblock Deep hough voting for 3d object detection in point clouds.
\newblock In \emph{Proceedings of the IEEE/CVF International Conference on Computer Vision}, 9277--9286.

\bibitem[{Radford et~al.(2021)Radford, Kim, Hallacy, Ramesh, Goh, Agarwal, Sastry, Askell, Mishkin, Clark et~al.}]{clip}
Radford, A.; Kim, J.~W.; Hallacy, C.; Ramesh, A.; Goh, G.; Agarwal, S.; Sastry, G.; Askell, A.; Mishkin, P.; Clark, J.; et~al. 2021.
\newblock Learning transferable visual models from natural language supervision.
\newblock In \emph{Proceedings of the International Conference on Machine Learning}, 8748--8763. PMLR.

\bibitem[{Rozenberszki, Litany, and Dai(2022)}]{scannet200}
Rozenberszki, D.; Litany, O.; and Dai, A. 2022.
\newblock Language-grounded indoor 3d semantic segmentation in the wild.
\newblock In \emph{Proceedings of the European Conference on Computer Vision}, 125--141.

\bibitem[{Song, Lichtenberg, and Xiao(2015)}]{song2015sunrgbd}
Song, S.; Lichtenberg, S.; and Xiao, J. 2015.
\newblock Sun RGB-D: A RGB-D scene understanding benchmark suite.
\newblock In \emph{Proceedings of the IEEE Conference on Computer Vision and Pattern Recognition}, 567--576.

\bibitem[{Vaswani et~al.(2017)Vaswani, Shazeer, Parmar, Uszkoreit, Jones, Gomez, Kaiser, and Polosukhin}]{vaswani2017attention}
Vaswani, A.; Shazeer, N.; Parmar, N.; Uszkoreit, J.; Jones, L.; Gomez, A.~N.; Kaiser, {\L}.; and Polosukhin, I. 2017.
\newblock Attention is all you need.
\newblock In \emph{Proceedings of the Advances in Neural Information Processing Systems}, volume~30.

\bibitem[{Wang et~al.(2023)Wang, Liu, Du, Ding, Liao, Qi, Chen, and Liu}]{wang2023object}
Wang, L.; Liu, Y.; Du, P.; Ding, Z.; Liao, Y.; Qi, Q.; Chen, B.; and Liu, S. 2023.
\newblock Object-Aware Distillation Pyramid for Open-Vocabulary Object Detection.
\newblock In \emph{Proceedings of the IEEE/CVF Conference on Computer Vision and Pattern Recognition}, 11186--11196.

\bibitem[{Xu et~al.(2023)Xu, Zhang, Wei, Hu, and Bai}]{SAN}
Xu, M.; Zhang, Z.; Wei, F.; Hu, H.; and Bai, X. 2023.
\newblock Side adapter network for open-vocabulary semantic segmentation.
\newblock In \emph{Proceedings of the IEEE/CVF Conference on Computer Vision and Pattern Recognition}, 2945--2954.

\bibitem[{Yang et~al.(2024)Yang, Liu, Lin, Hancke, and Lau}]{yang2024boosting}
Yang, Z.; Liu, Y.; Lin, J.; Hancke, G.; and Lau, R.~W. 2024.
\newblock Boosting weakly-supervised referring image segmentation via progressive comprehension.
\newblock \emph{arXiv preprint arXiv:2410.01544}.

\bibitem[{Zeng et~al.(2018)Zeng, Song, Welker, Lee, Rodriguez, and Funkhouser}]{robotic}
Zeng, A.; Song, S.; Welker, S.; Lee, J.; Rodriguez, A.; and Funkhouser, T. 2018.
\newblock Learning synergies between pushing and grasping with self-supervised deep reinforcement learning.
\newblock In \emph{Proceedings of the IEEE/RSJ International Conference on Intelligent Robots and Systems (IROS)}, 4238--4245.

\bibitem[{Zhang et~al.(2024)Zhang, Li, Zhang, Xie, Xue, Xie, and Zhang}]{fmov3d}
Zhang, D.; Li, C.; Zhang, R.; Xie, S.; Xue, W.; Xie, X.; and Zhang, S. 2024.
\newblock FM-OV3D: Foundation Model-Based Cross-Modal Knowledge Blending for Open-Vocabulary 3D Detection.
\newblock In \emph{Proceedings of the AAAI Conference on Artificial Intelligence}, volume~38, 16723--16731.

\bibitem[{Zhang et~al.(2022)Zhang, Guo, Zhang, Li, Miao, Cui, Qiao, Gao, and Li}]{zhang2022pointclip}
Zhang, R.; Guo, Z.; Zhang, W.; Li, K.; Miao, X.; Cui, B.; Qiao, Y.; Gao, P.; and Li, H. 2022.
\newblock Pointclip: Point cloud understanding by clip.
\newblock In \emph{Proceedings of the IEEE/CVF Conference on Computer Vision and Pattern Recognition}, 8552--8562.

\bibitem[{Zhang et~al.(2020)Zhang, Sun, Yang, and Huang}]{h3dnet}
Zhang, Z.; Sun, B.; Yang, H.; and Huang, Q. 2020.
\newblock H3dnet: 3d object detection using hybrid geometric primitives.
\newblock In \emph{Proceedings of the European Conference on Computer Vision}, 311--329.

\bibitem[{Zhou, Loy, and Dai(2022)}]{zhou2022extract}
Zhou, C.; Loy, C.~C.; and Dai, B. 2022.
\newblock Extract free dense labels from clip.
\newblock In \emph{Proceedings of the European Conference on Computer Vision}, 696--712.

\bibitem[{Zhou et~al.(2022)Zhou, Girdhar, Joulin, Kr{\"a}henb{\"u}hl, and Misra}]{3detic}
Zhou, X.; Girdhar, R.; Joulin, A.; Kr{\"a}henb{\"u}hl, P.; and Misra, I. 2022.
\newblock Detecting twenty-thousand classes using image-level supervision.
\newblock In \emph{Proceedings of the European Conference on Computer Vision}, 350--368.

\end{thebibliography}
\clearpage

\title{Hierarchical Cross-Modal Alignment for Open-Vocabulary 3D Object Detection
\\Supplementary Material}

\maketitle

\section{Analysis of Cross-Dataset Transferability}

\zyjnew{To verify the transferability of our method, we conduct two cross-dataset evaluations on ScanNet and SUN RGB-D datasets. In the first evaluation, we train our HCMA on ScanNet and test it on SUN RGB-D. Conversely, in the second evaluation, we train our HCMA on SUN RGB-D and test it on ScanNet. Results are reported in Table~\ref{cross_dataset}. Our HCMA demonstrates better performance compared to OV-3DET~\cite{ov-3det}. The results highlight the transferability of our HCMA, as it consistently outperforms the baseline method even when applied to datasets beyond its training dataset.}

\begin{table}[htb]
\centering
\renewcommand\arraystretch{1.4}
    \resizebox{0.415\textwidth}{!}
    {
    \begin{tabular}{cccc}
    \toprule
Training & Testing & Method & $mAP_{25}$\\
 \midrule
\multirow{2}{*}{ScanNet} & \multirow{2}{*}{SUN RGB-D} &OV-3DET & 12.31 \\
& & \textbf{HCMA} (Ours)	& \textbf{13.30}\\
 \midrule
\multirow{2}{*}{SUN RGB-D} & \multirow{2}{*}{ScanNet} &OV-3DET & 7.86  \\
& & \textbf{HCMA} (Ours)	& \textbf{8.94} \\
        \bottomrule

    \end{tabular}
    }    
    \caption{Results of cross-dataset evaluation on ScanNet and SUN RGB-D in terms of $mAP_{25}$.}
     \label{cross_dataset}   
\end{table}

\section{Analysis of Generalization Ability on \\ScanNet200 Dataset}
To verify the open-vocabulary capabilities of our method, we conduct evaluations on large-scale vocabularies using the ScanNet200 \cite{scannet200} dataset. The evaluation provides valuable insights into the robustness and scalability of our method, further validating its potential for practical applications in real-world 3D object detection scenarios.

Specifically, we train our HCMA on ScanNet and test it on the vocabularies in ScanNet200 without fine-tuning. 
Compared with our training vocabularies, ScanNet200 contains 53 overlapping vocabularies.
This indicates that there is a small degree of vocabulary overlap between our training data and the ScanNet200 dataset. Hence, it can be leveraged to verify the open-vocabulary ability of HCMA.

We compare our HCMA with OV-3DET \cite{ov-3det}, as shown in Table \ref{table_scannet200}. Our HCMA exhibits superior performance compared to OV-3DET \cite{ov-3det}, as measured by $mAP_{25}$ and $mAP_{50}$. This remarkable performance demonstrates the robustness and effectiveness of HCMA in handling large-scale open vocabularies.
This generalization ability is significant as it demonstrates the potential of our model in real-world applications.
\begin{table}[htb]
\centering
\renewcommand\arraystretch{1.4}
    \resizebox{0.3\textwidth}{!}
    {
    \begin{tabular}{ccc}
    \toprule
Method & $mAP_{25}$ &	$mAP_{50}$\\
 \midrule
OV-3DET & 2.39 & 0.84  \\
HCMA (Ours)	& \textbf{3.10} & \textbf{1.03} \\
        \bottomrule

    \end{tabular}
    }
    \caption{Result on ScanNet200 in terms of $mAP_{25}$ and $mAP_{50}$.}
    \label{table_scannet200}
\end{table}

\section{More Implementation Details}

We provide a comparison of computational overhead with baseline methods OV-3DET~\cite{ov-3det} and CoDA~\cite{cao2024coda} in Table~\ref{table_gpu}. We conduct our experiment on a single RTX4090 GPU, while OV-3DET and CoDA require 8 GPUs for the experiments. In addition, our training time is shorter than the baseline methods. These indicate that HCMA is not only more efficient in terms of computational resources but also faster in terms of model training.

We follow OV-3DET~\cite{ov-3det} to present the baseline results of OV-PointCLIP~\cite{zhang2022pointclip}, OV-3DETIC~\cite{3detic}, and OV-CLIP-3D~\cite{clip}:

\noindent\textbf{OV-PointCLIP.} We first employ pre-trained 3DETR~\cite{3detr} for 3D detection. Following this step, points are extracted inside the predicted 3D bounding boxes. Subsequently, the pre-trained PointCLIP~\cite{zhang2022pointclip} is utilized for object classification based on these points, facilitating the assignment of predicted classes to each predicted 3D bounding box.

\noindent\textbf{OV-3DETIC.} We begin by utilizing Detic~\cite{3detic} to predict 2D bounding boxes (Eq.~\ref{eq1}) of the main paper and classes on images. We then back-project these 2D bounding boxes into 3D space as 3D bounding boxes (Eq.~\ref{eq2}) of the main paper.

\noindent\textbf{OV-CLIP-3D.} Similar to OV-3DETIC, which also utilizes Detic~\cite{3detic} to predict 2D bounding boxes (Eq.~\ref{eq1}) of the main paper and transfer them into 3D bounding boxes (Eq.~\ref{eq2}) of the main paper. However, it diverges by employing CLIP~\cite{clip} to perform open-vocabulary classification instead of class prediction from Detic~\cite{3detic}.

\begin{table}[htb]
\centering
    \resizebox{0.45\textwidth}{!}
    {
    \begin{tabular}{cccc}
    \toprule
\multirow{2}{*}{Method} & Computational &	\multicolumn{2}{c}{Training Time}\\
    & Resource & Phase 1 & Phase 2 \\
 \midrule
OV-3DET & 8$\times$2080Ti & 48 hours & 24 hours \\
CoDA	& 8$\times$V100 & 2-3 days & 1-2 days \\
HCMA(Ours) & 1$\times$4090 & 40 hours & 20 hours \\
        \bottomrule

    \end{tabular}
    }
    \caption{Comparison of computational overhead.}
    \label{table_gpu}
\end{table}

\section{More Quantitative Comparison}
To compare our HCMA with the latest OV-3DOD method, FM-OV3D~\cite{fmov3d}, 
we adapt HCMA following the experimental setting of FM-OV3D~\cite{fmov3d}. In this evaluation, we compare HCMA with OV-3DIC~\cite{ov-3detic} and FM-OV3D~\cite{fmov3d} on  ScanNet~\cite{dai2017scannet} and SUN RGB-D~\cite{song2015sunrgbd} datasets. FM-OV3D* denotes the model trained in an
annotation-free setting, and FM-OV3D represents the model trained only utilizing knowledge blending. As Tables~\ref{more_scannet} and Table~\ref{more_sunrgbd} show, our HCMA outperforms the latest OV-3DOD method on both ScanNet and SUN RGB-D datasets by a large margin, demonstrating the superior performance of HCMA.
\begin{table*}[htb]\centering
    \resizebox{\textwidth}{!}{
    \large
    \begin{tabular}{c|c|cccccccccc}
    \toprule
        Method & $mAP_{25}$ & toilet & bed & chair & sofa & dresser & table & cabinet & bookshelf & pillow & sink\\
    \midrule
    OV-3DIC~\cite{ov-3detic} & 12.65 & 48.99 &2.63& 7.27 &18.64 &2.77& 14.34& 2.35 &4.54& 3.93 &21.08 \\
FM-OV3D* \cite{fmov3d} & 14.34 &2.17 &41.11 &27.91& 33.25 &0.67 &12.60& 2.28 &8.47& 9.08 &5.83 \\
FM-OV3D \cite{fmov3d} &21.53 &  62.32& 41.97& 22.24 &31.80 &1.89& 10.73& 1.38& 0.11& 12.26 &30.62 \\
\midrule
        \textbf{HCMA} (Ours) & \textbf{31.63} & \textbf{72.88} & \textbf{50.64} & \textbf{37.28} & \textbf{56.83} & \textbf{3.11} & \textbf{15.30} & \textbf{3.10} & \textbf{11.79} & \textbf{20.91} & \textbf{44.49} \\
        \bottomrule
    \end{tabular}
    }
    \caption{\zyjnew{Results on ScanNet in terms of $AP_{25}$. We report \jy{the average} value of all 10 categories.}}
    \label{more_scannet}
\end{table*}
\begin{table*}[htb]\centering
    \resizebox{\textwidth}{!}{
    \large
    \begin{tabular}{c|c|cccccccccccccccccccc}
    \toprule
        Method & $mAP_{25}$ & toilet & bed & chair & bathtub & sofa & dresser & scanner & fridge & lamp & desk \\
    \midrule
    OV-3DIC~\cite{ov-3detic} & 13.03 & 43.97& 6.17& 0.89 &45.75& 2.26& \textbf{8.22} &0.02& 8.32& 0.07& \textbf{14.60} \\
FM-OV3D* \cite{fmov3d} & 16.98 & 32.40 &18.81& 27.82& 15.14 & 35.40& 7.53& \textbf{1.95} & 9.67& 13.57& 7.47 \\
FM-OV3D \cite{fmov3d} &21.47 &  55.00 &38.80 &19.20 &41.91& 23.82& 3.52& 0.36& 5.95 &\textbf{17.40}& 8.77 \\
\midrule
        \textbf{HCMA} (Ours) & \textbf{32.35} & \textbf{68.81} & \textbf{72.87} & \textbf{41.63} & \textbf{49.90} & \textbf{43.18} & 3.62 & 0.05 & \textbf{16.71} & 14.52 & 12.26 \\
   \bottomrule
    \end{tabular}
    }
    \caption{\zyjnew{Results on SUN RGB-D in terms of $AP_{25}$. We report \jy{the average} value of all 10 categories.}}
    \label{more_sunrgbd}
\end{table*}

\section{Implementation of the Alignment Loss Function}
Table~\ref{align_loss} shows the implementation of the alignment loss function in Eq. 11. In this function, $\mathds{1}$ is the indicator function that controls the specific form of $\mathcal{L}_{align}$, which is decided by different hierarchies utilized in the training stage.
\begin{table}[htb]
\centering
\renewcommand\arraystretch{1.4}
    \resizebox{0.4\textwidth}{!}
    {
    \begin{tabular}{ccccccc}
        \toprule
       HDI & $\mathcal{L}_{m}^{o}$ & $\mathcal{L}_{m}^{v}$ & $\mathcal{L}_{m}^{s}$ & $\mathcal{L}_{m}^{l}$ & $\mathcal{L}_{m}^{g}$ & $\mathcal{L}_{m}^{a}$ \\
        \midrule
        O &  \checkmark &  &  &   \\
        O V &  &  \checkmark &  &  \checkmark  \\
        O S &   &  &  \checkmark &   &  \checkmark &    \\
        O V S &   & \checkmark &  \checkmark &   &   & \checkmark \\
        \bottomrule
    \end{tabular}
    }
    \caption{Components of the alignment loss. O, V, and S denote object, view, and scene levels, respectively.}
    \label{align_loss}    
\end{table}

\section{Vocabulary Details}
In our proposed HCMA, the use of vocabulary can be divided into two stages, including 1) pseudo-3D bounding box generation and cross-modal alignment in the training stage, and 2) object detection in the testing stage. Specifically, we use categories from LVIS \cite{gupta2019lvis} as training vocabularies in the training stage. The baseline methods use the same vocabulary set during training for the fair comparison.
\begin{table}[t]
\centering
\renewcommand\arraystretch{1.4}
    \resizebox{0.4\textwidth}{!}
    {
    \begin{tabular}{lccc}
        \toprule
       Testing Benchmark & Total & Overlap & Open\\
        \midrule
        ScanNet  & 20 & 12 & 8\\
        ScanNet200  & 200 & 53 & 147\\ 
        SUN RGB-D  & 20 & 14 & 6\\
        \bottomrule
    \end{tabular}
    }
    \caption{Relationship between training and testing vocabularies.}
    \label{vocabulary_overlap}  
\end{table}

In the testing stage, we sample 20 common categories as testing vocabularies in the evaluation on the ScanNet \cite{dai2017scannet} and SUN RGB-D \cite{song2015sunrgbd} benchmarks, respectively. As for ScanNet200 \cite{scannet200}, we follow the default vocabulary setting in this benchmark.
The relationship of the vocabulary set between training and the testing phase is shown in Table \ref{vocabulary_overlap}. Open vocabulary refers to vocabularies used in the testing phase but not in the training phase, while overlapping vocabulary includes vocabularies that are both used in the training and testing phases.

\section{Results on Open Vocabulary}

To further verify the open-vocabulary ability of our HCMA, we conduct additional experiments involving 3D object detection on the ScanNet~\cite{dai2017scannet} and SUN RGB-D~\cite{song2015sunrgbd} datasets, focusing solely on open vocabularies. The number of open vocabularies used in this experiment is presented in Table~\ref{vocabulary_overlap}. The evaluation results, shown in Table \ref{open_voca_map25}, indicate that HCMA outperforms the baseline method OV-3DET \cite{ov-3det}. It demonstrates the capabilities of HCMA on open-vocabulary 3D object detection. However, we observe that there is still a large gap in 3D object detection results between the open-vocabulary and the seen-vocabulary. This observation demonstrates the significant potential for further development in 3D object detection involving open-vocabulary scenarios.

\begin{table}[h]
\centering
\renewcommand\arraystretch{1.4}
    \resizebox{0.35\textwidth}{!}
    {
    \begin{tabular}{ccc}
    \toprule
Method & ScanNet &	SUN RGB-D \\
 \midrule
OV-3DET & 7.80 & 6.29  \\
HCMA (Ours)	& \textbf{8.09} & \textbf{6.60} \\
        \bottomrule

    \end{tabular}
    }
    \caption{Results of open vocabularies on ScanNet and SUN RGB-D in terms of $mAP_{25}$.}
    \label{open_voca_map25}
\end{table}

\begin{table*}[htb]\centering
    \resizebox{\textwidth}{!}{
    \large
    \begin{tabular}{c|c|cccccccccccccccccccc}
    \toprule
        Method & $mAP_{25}$ & \rotatebox{90}{toilet} & \rotatebox{90}{bed} & \rotatebox{90}{chair} & \rotatebox{90}{sofa} & \rotatebox{90}{dresser} & \rotatebox{90}{table} & \rotatebox{90}{cabinet} & \rotatebox{90}{bookshelf} & \rotatebox{90}{pillow} & \rotatebox{90}{sink} & \rotatebox{90}{bathtub} & \rotatebox{90}{refrigerator} & \rotatebox{90}{desk} & \rotatebox{90}{night stand} & \rotatebox{90}{counter} & \rotatebox{90}{door} & \rotatebox{90}{curtain} & \rotatebox{90}{box} & \rotatebox{90}{lamp} & \rotatebox{90}{bag}\\
    \midrule
3DETR & 46.95 & 90.75 & 62.97 & 67.56 & 68.42 & 34.29 & 53.48 & 38.90 & 44.22 & 43.66 & 62.56 & 77.27 & 43.67 & 52.54 & 50.61 & 28.15 & 44.63 & 36.53 & 11.96 & 17.66 & 9.13 \\
        \textbf{HCMA} (Ours) & 21.77 & 72.85 & 50.61 & 37.26 &56.82 & 3.11 & 15.19 & 3.10 & 11.78 & 20.89 & 44.51 & 50.13 & 9.49 & 12.73 & 24.59 & 0.10 & 13.56 & 0.19 & 3.27 & 3.28 & 1.86 \\
        \bottomrule
    \end{tabular}
    }
    \caption{\zyjnew{Results on ScanNet in terms of $AP_{25}$. We report \jy{the average} value of all 20 categories.}}
    \label{upperbound_scannet}
\end{table*}
\begin{table*}[htb]\centering
    \resizebox{\textwidth}{!}{
    \large
    \begin{tabular}{c|c|cccccccccccccccccccc}
    \toprule
        Method & $mAP_{25}$ & \rotatebox{90}{toilet} & \rotatebox{90}{bed} & \rotatebox{90}{chair} & \rotatebox{90}{bathtub} & \rotatebox{90}{sofa} & \rotatebox{90}{dresser} & \rotatebox{90}{scanner} & \rotatebox{90}{fridge} & \rotatebox{90}{lamp} & \rotatebox{90}{desk} & \rotatebox{90}{table} & \rotatebox{90}{stand} & \rotatebox{90}{cabinet} & \rotatebox{90}{counter} & \rotatebox{90}{bin} & \rotatebox{90}{bookshelf} & \rotatebox{90}{pillow} & \rotatebox{90}{microwave} & \rotatebox{90}{sink} & \rotatebox{90}{stool}\\
    \midrule
    3DETR &39.50& 89.62& 82.08& 65.91& 74.20& 57.06 &24.49& 12.49& 24.43& 24.94 &28.17& 49.74& 59.71& 18.18& 28.86& 43.76 &31.58& 19.45 &10.07 &31.61 &13.55\\
    \textbf{HCMA} (ours) &21.53 & 68.50 & 72.81 & 40.59 & 49.65 & 43.20 & 3.32 & 0.03 & 17.38 & 14.34 & 11.73 & 28.61 & 19.48 & 0.56 & 0.12 & 11.33 & 1.51 & 10.34 & 1.34 & 31.56 & 4.10 \\
   \bottomrule
    \end{tabular}
    }
    \caption{\zyjnew{Results on SUN RGB-D in terms of $AP_{25}$. We report \jy{the average} value of all 20 categories.}}
    \label{upperbound_sunrgbd}
\end{table*}

\section{Analysis of the Upper Bound Performance}
Since we utilize 3DETR \cite{3detr} as our 3D detector, we use 3DETR as our upper bound. 3DETR \cite{3detr} is trained with ground truth 3D annotations in a fully supervised manner, while HCMA performs OV-3DOD without 3D annotations. Results are shown in Tables~\ref{upperbound_scannet} and~\ref{upperbound_sunrgbd}. It indicates that there is still a large gap between fully-supervised 3D detection and open-vocabulary 3D detection. While advancements have been made in the field of 2D open-vocabulary understanding, achieving comparable performance to fully-supervised methods remains a challenging task in the 3D domain. 

\section{Analysis of the Top-Down Image}
The top-down image can encompass the majority of objects in the 3D scene, providing a comprehensive view of the spatial relations and context among all objects. Especially when objects are distantly located, the top-down image can directly capture their structural information, but multi-view images struggle to achieve this. While multi-view images can provide local information, they may lack the broader global context that the top-down image can provide. These deficiencies can be effectively addressed by incorporating the top-down image to offer global information. We emphasize the significance of this comprehensive 3D scene context in the 3D scene understanding task during the feature alignment in our ICMA module.

We conduct an experiment that removes the top-down image from the ICMA module in the ScanNet~\cite{scannet200} dataset. In this setting, both the intra-level alignment between scene-level image and scene-level point cloud, as well as the inter-level alignment between scene-level image and view-level image, are removed. As shown in Table~\ref{topdown}, the performance improves when top-down images are introduced, which suggests that top-down images can enhance 3D scene understanding through intra-level and inter-level alignment in our ICMA module.

\begin{table}[h]
\centering
\renewcommand\arraystretch{1.4}
    \resizebox{0.3\textwidth}{!}
    {
    \begin{tabular}{ccc}
    \toprule
Top-Down Image & $mAP_{25}$ &	$mAP_{50}$ \\
 \midrule
 & 20.21 &	6.23  \\
	\checkmark & \textbf{20.30}  &	\textbf{6.39} \\
        \bottomrule

    \end{tabular}
    }
    \caption{Ablation study of the top-down image.}
    \label{topdown}
\end{table}

\section{Analysis of the Scene Label}
We conduct an ablation study on the use of scene labels in generating the positive and negative samples. As shown in Table~\ref{scenelabel}, the results exhibit improvement when employing scene labels for alignment, indicating that using scene labels to construct view-level pairs is effective.

\begin{table}[t]
\centering
\renewcommand\arraystretch{1.4}
    \resizebox{0.3\textwidth}{!}
    {
    \begin{tabular}{ccc}
    \toprule
Scene Label & $mAP_{25}$ &	$mAP_{50}$ \\
 \midrule
 & 18.89	& 5.74  \\
	\checkmark & \textbf{19.10} &	\textbf{5.84} \\
        \bottomrule

    \end{tabular}
    }
    \caption{Ablation study of the scene label.}
    \label{scenelabel}
\end{table}

\begin{figure*}[ht] \centering

    \includegraphics[width=\textwidth]{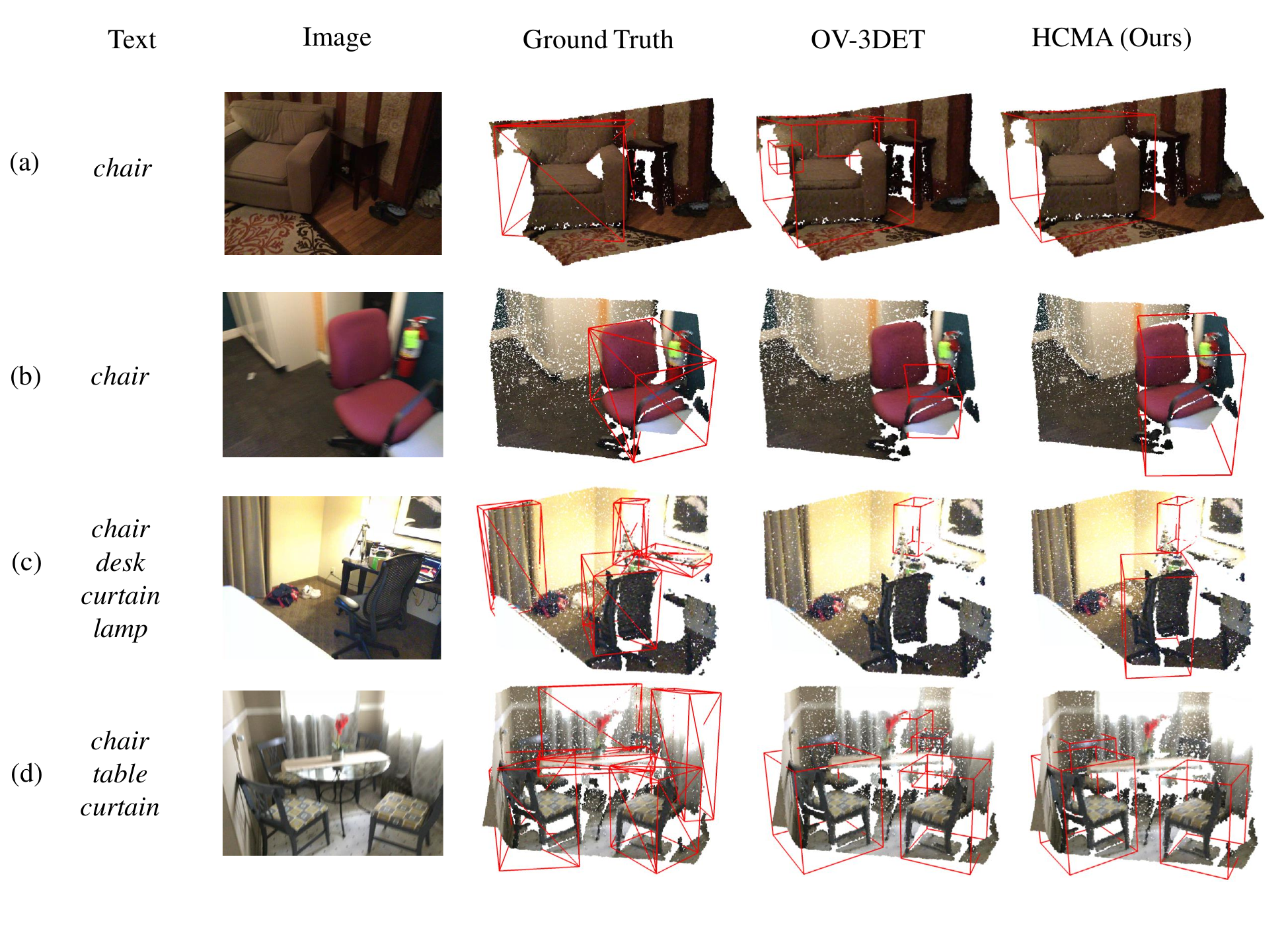}
    \vspace{-4mm}
    \caption{More qualitative comparisons with OV-3DET \cite{ov-3det}. For each case, the detection text prompts are shown on the left. Note that HCMA only utilizes coordinate information of the 3D point cloud and does not include color information. For better visualization, the colors of the 3D point clouds are displayed.} 
    \label{visual_img_supp}
    \vspace{-4mm}
\end{figure*}

\section{More Qualitative Results}

In this section, we provide more qualitative results on ScanNet~\cite{dai2017scannet} in Figure \ref{visual_img_supp}. It shows that our HCMA can correctly predict the location, size, and orientation of the 3D bounding box. In contrast, the baseline OV-3DET \cite{ov-3det} may exhibit inaccuracies in predicting the size and number of 3D bounding boxes, such as the chair in sample $a$ and sample $b$. The baseline OV-3DET may miss the object in prediction. For example, the chair in sample $c$ is an example of a target object missed by the baseline method, while HCMA can detect the chair in the 3D scene.
The observed discrepancy between HCMA and OV-3DET can be attributed to the differences in their ability to comprehend the overall structure of the target objects. OV-3DET may struggle to understand the holistic information of the objects, leading to the generation of small and redundant 3D bounding boxes. In contrast, HCMA leverages scene context from diverse hierarchies, enabling a better understanding of the comprehensive information regarding the 3D objects. This enhanced contextual understanding allows HCMA to generate more accurate and correct 3D bounding boxes.
\subsection{Failure Cases}
We observe that HCMA may sometimes miss the target object in the 3D scene. For example, both HCMA and OV-3DET cannot detect the curtain and table in sample $d$. The transparent glass table makes it very difficult to be detected since point clouds often struggle to accurately represent transparent objects. Additionally, detecting objects under open vocabularies, such as curtains, can also present difficulties. Open-vocabulary objects may not be sufficiently presented in the training data, resulting in limited recognition capabilities for these objects during testing.

\section{Limitations}
\jynew{Our proposed HCMA demonstrates promising results in the OV-3DOD task. However, it still inherits the limitations of current open-vocabulary 3D methods, such as the requirement for pre-defined vocabularies. \rynn{As a future work, we would like to explore} designing an OV-3DOD method that does not depend on pre-defined vocabularies.}

\section{Broader Impacts}
Our paper does not have direct societal impact. We train our method on public databases, with no private data used. While we do not foresee any negative societal impact from our paper, it may be leveraged in personal 3D data, resulting in leakage risks that raise privacy concerns. We urge readers to limit the usage of this work to legal use cases.

\end{document}